\documentclass{llncs}
\usepackage{graphicx}
\usepackage{amsmath}
\usepackage[export]{adjustbox}
\begin{document}

\title{FaceOff: Anonymizing Videos in the Operating Rooms}
\titlerunning{Faster-RC3NN}  
%
%
\author{Evangello Flouty\inst{1} \and Odysseas Zisimopoulos\inst{1} \and Danail Stoyanov\inst{1,2}}
\authorrunning{} 
%
%
\institute{Digital Surgery, London, United Kingdom\\
\and
Wellcome / ESPRC Centre for Interventional and Surgical Sciences, London, United Kingdom\\
}

\maketitle              

\begin{abstract}
Video capture in the surgical operating room (OR) is increasingly possible and has potential for use with computer assisted interventions (CAI), surgical data science and within smart OR integration. Captured video innately carries sensitive information that should not be completely visible in order to preserve the patient's and the clinical teams' identities. When surgical video streams are stored on a server, the videos must be anonymized prior to storage if taken outside of the hospital. In this article, we describe how a deep learning model, Faster R-CNN, can be used for this purpose and help to anonymize video data captured in the OR. The model detects and blurs faces in an effort to preserve anonymity. After testing an existing face detection trained model, a new dataset tailored to the surgical environment, with faces obstructed by surgical masks and caps, was collected for fine-tuning to achieve higher face-detection rates in the OR. We also propose a temporal regularisation kernel to improve recall rates. The fine-tuned model achieves a face detection recall of 88.05\% and 93.45 \% before and after applying temporal-smoothing respectively.
\end{abstract}

\section{Introduction}
Video cameras are pervasive within the modern operating room (OR) and used extensively during surgery, for example in laparoscopic or robotic assisted surgery, but with minimal video utilization. Specifically many integrated operating rooms now incorporate surveillance cameras or documentation cameras integrated within the surgical lights or in the ceiling. The video data collected by such devices is highly sensitive because it records events during the operation and also the identities of staff and patients within the OR. Yet, the video can have multiple uses in educational material or in the analysis and automation of OR optimisation systems through surgical data science platforms\cite{sds}. To be able to use the recorded videos in the OR, video processing must take place to ensure the data is anonymized and safe to be used. It is possible to approach video anonymization through computer vision algorithms for face detection but making such systems work well in surgical environments is difficult because the OR has variable lighting conditions, multiple occlusion possibilities and also the team wears surgical drapes and masks.

\begin{figure}[t!]
\centering
\includegraphics[width=1\columnwidth]{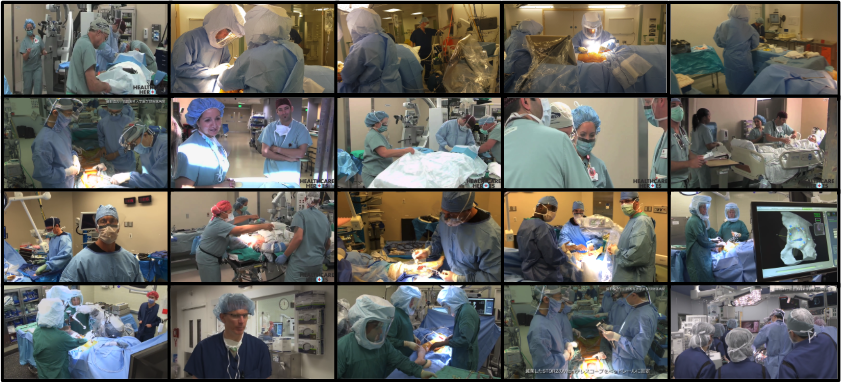}
\caption{FaceOff images, collected from Youtube, showing the faces in the surgical environment potentially exposing sensitive information}
\label{videos}
\end{figure}

Real-time face detection is a mature field in computer vision \cite{VJ2014}. As with many problems in the field, techniques using hand crafted features such as HOG\cite{dalalN}, have recently been superseded by convolutional neural networks (CNNs) based approaches using deep learning for detection\cite{fdl1}\cite{fdl2}\cite{fdl3}, pose estimation\cite{pe1}, and emotion prediction\cite{ed1}. The introduction of big datasets such as FDDB\cite{fddb}, IJB-A\cite{ijba}, and WIDER\cite{wider} has empowered the use of deep learning models and enhanced robustness and efficiency, shown by the evolution of approaches from recurrent CNN (RCNN) \cite{rcnn}, followed by Fast-RCNN\cite{fastrcnn}, and finally Faster-RCNN\cite{fasterrcnn}. The results for these architectures are impressive but their translation into the clinical setting faces challenges because the data needs adaptation to deal with masked faces, surgical caps and the lighting variability within the room.

In this paper, we adopt the Faster-RCNN model pre-trained on the available WIDER dataset and we adapt it for face detection in the OR. Faces in the OR are very different from the WIDER dataset due to masks, caps, and surgical magnifying glasses. Detecting such faces is difficult and requires model adaptation, which we achieve through collecting surgical data from web search engines, labelled and used to fine-tune the model. To achieve anonymization, it is important that the model catches as many faces as possible. A sliding window for temporal smoothing was implemented and then applied on the detections to have a higher chance of detecting any missed face (a false negative). Our method shows promising results on our validation dataset which will be made available to the community.

\section{Methods and Data}

\noindent\textbf{Wider Dataset}
The dataset consists of 32,203 images with 393,703 faces in 61 different environments (meetings, concerts, parades, etc \dots). It is also worth noting that this dataset include 166 images (in the training set) of faces in the surgical environment. This dataset is commonly used for benchmarking face detection. Faster RCNN is in the top 4 of all the submissions that used the WIDER dataset to benchmark performance\cite{facefaster}.\\ 
\noindent\textbf{FaceOff Dataset}
We collected 15 videos of surgical ORs from the video search engine Youtube. All were publicly available with "Standard Youtube License" (videos can be used freely). The keywords used for searching: surgery, realtime surgery, surgery in the operating room/theatre, recorded surgery... Figure \ref{videos} shows a sample of the dataset. In total, the dataset consists of $6371$ images describing $12786$ faces. The images show variability in scales and occlusions of faces in the OR to achieve a good learning of the facial features in the OR.

\subsection{Faster R-CNN}
Faster R-CNN uses a regional proposal network (RPN) that estimates bounding boxes around regions in the input image. It is scale invariant as it proposes regions of many scales before interrogating each with one of two CNNs: ZFnet\cite{zfnet} and VGG-16\cite{vgg16}. The convolutional layers are shared with the RPN (unlike the architecture in Fast R-CNN), making computation efficient. The CNNs evaluate regions using the intersection of union (IoU) of each anchor with the ground truth bounding boxes of the input image during training to determine if the region is used as a positive or negative sample. The RPN proposes around $21000$ regions per image but after non-max filtering (NMF) around $2000$ valid anchors remain and only $256$ positive anchors, and $256$ negative anchors are then chosen for training.\\

The loss function of the RPN incorporates several parts shown in the equations below:

\begin{equation}
\mathcal{L}(p_i, t_i)= \dfrac{1}{N_{cls}}\sum\limits_{i} \mathcal{L}_{cls}(p_i,p_i^*) + \lambda \dfrac{1}{N_{reg}}\sum\limits_{i} p_i^*\mathcal{L}_{reg}(t_i,t_i^*)
\label{loss}
\end{equation}

\begin{equation}
\mathcal{L}_{cls}(p_i,p_i^*) = -log(\dfrac{e^{f_{y_i}}}{\sum\limits{j}e^{f_{j}}})
\label{losspart1}
\end{equation}
\begin{equation}
\mathcal{L}_{reg}(t_i,t_i^*) = smooth_{L_1}(t_i-t_i^*)=\begin{cases}
0.5\text{ } (t_i-t_i^*)^2,& \text{if $|t_i-t_i^*|<1$}\\
|t_i-t_i^*| - 0.5,& \text{otherwise}
\end{cases}
\label{losspart2}
\end{equation}

The first part measures the error of the classifier whether the region is a class (in this case a face) or not. Where $p_i$ is the predicted probability, $p_i^*$ is either $0$ (when the region describes the background class) or $1$ (when the region describes the foreground class, in this case a face), and finally $N_{cls}$ is the mini-batch size (in this case 2*256 = 512). The classifier loss as shown in equation (\ref{losspart1}) is the soft-max loss of the predicted class. The second part tries to measure the error of box regressors. Where $\lambda$ is a constant, $p_i^*$ is the predicted probability (this means this part of the equation is only activated for positive anchors where $p_i^*$=1),$t_i$ is the predicted box, $t_i^*$ the ground truth box, and finally $N_{reg}$ is the total number of valid anchors (in this case around $2000$). The box regressor loss is the smoothing function of the predicted box as shown in equation (\ref{losspart2}). It tries to minimize the difference between the predicted box and the ground truth box.

\begin{figure}[t!]
\includegraphics[width=1\columnwidth]{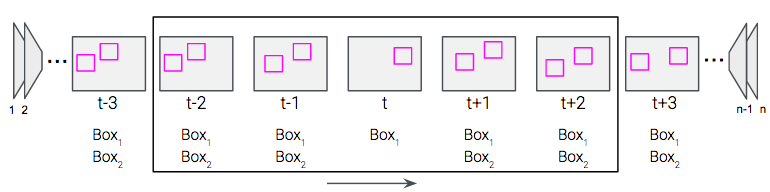}
\caption{The kernel size of the sliding average window  is 5. The sliding average kernel estimates the missing $Box_2$ at frame $t$ by averaging the corresponding detected $Box_2$ in the adjacent frames $\in [t-2, t+2]$. }
\label{smoother}
\end{figure}

\subsection{Sliding Kernel Smoother}
Despite the excellent performance of a per-frame face detection method, temporal discontinuities are still possible and need to be handled with a non-detection driven approach. For anonymization, having a high recall (or low false negatives) is the main target for the model to achieve. While the described Faster R-CNN captures spatial information exceptionally well it can suffer from period occlusion or failure when faces turn or enter variable illumination conditions. But since videos will be inferred using the model, valuable temporal information can potentially be lost. As illustrated in the Fig 2 schematic, the model sometimes misses faces even though it successfully detected the same face in adjacent frames. To take advantage of that, a sliding window of kernel sizes $k=3, 5, 7$ were applied to smooth in the detections to be able to anonymize the missed faces. Doing so will also generate more false positives as the smoothing kernel does not incorporate visual information. As described in Fig 2, the smoothing window will apply a moving average on the centre frame $t$ and estimates $Box_2$ at frame $t$ with the aim of anonymizing a missed face.


%
\begin{figure}[t!]
\begin{center}
\includegraphics[width=0.95\columnwidth]{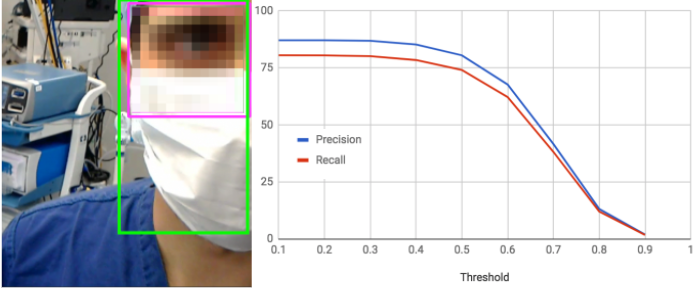}
\label{threshold}
\end{center}
\caption{\textbf{ Left}: An example case showing the intuition of picking the right IOU threshold to calculate the metrics. The green and pink bounding boxes describes the ground truth and detected face respectively. As seen in the image, the anonymization has occurred given the area above the mask was detected. The detected region is less than half the area of the annotated face. Therefore, a threshold of $t = 0.3$ was used. \textbf{Right}: The precision and recall as a function of the value of the IOU threshold that counts a detection whether its a true positive or a false positive}
\end{figure}

\section{Experiments and Results}
\subsubsection{Calculating activations}: Given that the model returns a bounding box, a metric must quantify how correct is that bounding box. This section will explain how those metrics where calculated. There are 4 detection cases that occur after inferring the test set. The first case occurs when the intersection over union (IOU) between the ground truth box and the predicted box is above a certain threshold $t$. This detection counts as a true positive. The second case occurs when there is no detected box close to a ground truth box. This counts as a false negative. The third case happens when there is a detected box without a ground truth box around it. This case counts as a false positive. Finally in the fourth case, when the IOU of the ground truth with the detected bounding box is lower that a threshold $t$, it counts as a false positive and a false negative (one for missing the detection, and one for detecting something that is not a face).

To set the threshold $t$, the precision and recall were calculated for 9 possible values. The results can be seen in the right section of figure 3. Intuitively speaking, both the precision and recall will drop as the IOU threshold increase as it will be less likely for the predicted box to be more aligned with the ground truth. This graph shows that the precision and recall are stable between $0.1$ and $0.3$. They start slowly decreasing between $t \in [0.4,0.5]$. A sharp drop is observed after $0.5$. After evaluating the above graph, a threshold of $t = 0.3$ was chosen. $0.3$ is a good value for the IOU threshold because faces are mostly covered with surgical masks. The detections sometimes only cover the eye area as shown in the left section of figure 3, even thought the ground truth describes the whole face including the mask. This is a good detection as it anonymizes the face and therefore it must be counted as a true positive.

\noindent\textbf{WIDER fine-tuning setup}: For a better anonymization, detecting normal faces is also crucial in the operating room. For that, the model from \cite{facefaster} was used. This paper fine-tuned a VGG-16 faster r-cnn trained on Imagenet using the WIDER dataset. They used stochastic gradient descent (SGD) for $50000$ iterations with base learning rate of $10^{-3}$ and then ran another $30000$ iterations with a base learning rate of $10^{-4}$.
\begin{figure}[t!]
\includegraphics[width=0.8\columnwidth,center]{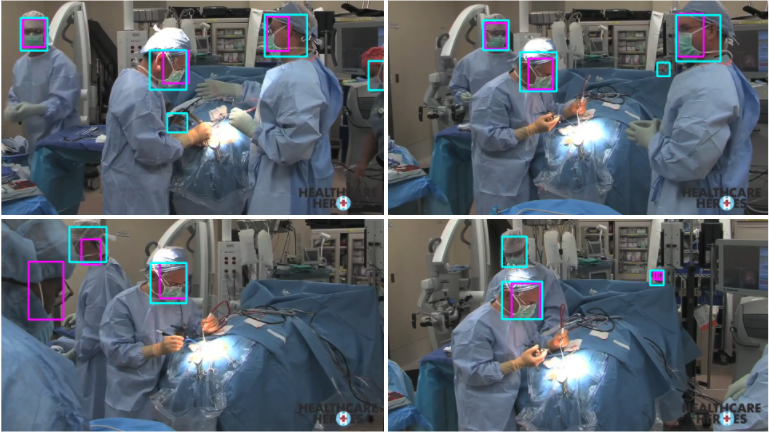}
\caption{Sample detections of both models. The WIDER trained model detections are shown in pink and the FaceOff fine-tuned model detections are shown in blue.}
\label{detections}
\end{figure}

\noindent\textbf{FaceOff fine-tuning setup}: After training the model described above, we further fine-tuned the model on the newly collected dataset of faces in the OR. We trained the model on $8485$ faces in the OR for $20000$ iterations. The RPN generates 12000 and 2000 ROIs before and after applying NMF respectively. Model uses a mini-batch (batch of regions of ROIs) size of 64 (for normalisation), an IOU threshold of 0.7 and above to consider the ROI as an example of a face, and an IOU threshold of 0.3 and less to consider the ROI as an example of a background. The remaining ROIs (with IOU between $[0.3,0.7]$ are discarded). Finally, a size set of 256 regions per class (256 regions for the face class, and 256 regions for the background class) is used for training.

We inferred the test set using the model trained on the WIDER dataset. The model returned a precision of 66.84\%, a recall of 75.40\%, and an F1 score of 70.86\%. After those promising results, we fine-tuned the model using the FACEOFF collected dataset with the setup discussed above. A precision of 82.58\%, recall of 88.05\%, and and f1 score of 85.23\% was achieved. A sample of the detections can be seen in figure \ref{detections}.

In the surgical environment, the model must achieve a high recall since it is more important to detect a face than to falsely detect a face. In other words, the volume of false negatives should be as small as possible irrespective of the volume of false positives. To take advantage of the temporal information found in a video, the detections where smoothed around frames with no detections. Surrounding frames have very similar information with a high probability. Averaging the surrounding detections around a frame should help in detect false negatives. The disadvantage of this approach is that it is more likely to generate false positive than detecting false negatives. After getting the detections from the FaceOff fine-tuned Faster R-CNN model, a sliding window of kernel $ k =3,5,7$ was explored. Table \ref{results} shows that the kernel of size 3 performed the best achieving a recall of 93.46 \%.

\begin{center}
\begin{table}[t]
\renewcommand{\arraystretch}{1.5}
\renewcommand{\tabcolsep}{5mm}
\caption{Surgical face detection metrics of the different models tested.}
\centering
\scalebox{0.95}{
\begin{tabular}  {|c | c | c | c | }
\hline
 Model & Precision & Recall & F1\\
\hline
Off-the-shelf & 66.84\% & 75.40\$ & 70.86\% \\
\hline
Fine-Tuned on FaceOff & 82.58\% &88.05\%&85.23\% \\
\hline
Post-Smoothing k = 3 & 59.07\% & \textbf{93.46}\% & 72.39\% \\
\hline
Post-Smoothing k = 5 & 55.93\% & 93.45\% & 69.96\% \\
\hline
Post-Smoothing k = 7 & 53.52\% & 93.26\% & 68.01\% \\
\hline
\end{tabular}}
\label{results}
\end{table}
\end{center}

\section{Discussion and Conclusion}
An increasing number of cameras are integrated in the OR (head mounted, ceiling mounted, light integrated, etc.) and anonymization of video is important in order to be able to use the recorded data for a wide range of purposes like documentation, teaching and surgical data science. In order to automatically blur faces in the recorded video, we have described a method and dataset that adapts the state-of-the-art face detection techniques. Our FaceOff method and dataset describe faces in the surgical environment and use temporal smoothing to increase the recall of detection and hence increase the effectiveness of video anonymization. We fine-tuned the Faster R-CNN pretrained on the face-detection-benchmark WIDER dataset achieving a recall of 88.05 \%. Taking advantage of the temporal nature of the application (anonymizing surgical video), a sliding average window was applied to the detections to smooth the missed detected faces reaching a recall of 93.46\% on the collected FaceOff test-set. The work described in our study is a first step towards building the tools and capabilities needed in order to begin taking advantage of surgical data and building surgical data science pipelines.

\section*{Acknowledgements}
We gratefully acknowledge the work and support received from the Innovation team at Digital Surgery.\\
Danail Stoyanov receives funding from the EPSRC (EP/N013220/1, EP/N022750/1, EP/N027078/1, NS/A000027/1), Wellcome/EPSRC Centre for Interventional and Surgical Sciences (WEISS) (203145Z/16/Z) and EU-Horizon2020 (H2020-ICT-2015-688592).
%
%


\begin{thebibliography}{5}
%
\bibitem {sds}
Maier-Hein, L.,\textit{et al.}, 2017. Surgical data science: enabling next-generation surgery. arXiv preprint arXiv:1701.06482.

\bibitem {VJ2014}
Viola, P. and Jones, M.J., \textit{2004}. Robust real-time face detection. International journal of computer vision, 57(2), pp.137-154.
\bibitem {dalalN}
Dalal, N. and Triggs, B., \textit{2005}, June. Histograms of oriented gradients for human detection. In Computer Vision and Pattern Recognition, 2005. CVPR 2005. IEEE Computer Society Conference on (Vol. 1, pp. 886-893). IEEE.
\bibitem {fdl1}
Liu, Z., \textit{et al.}, 2015. Deep learning face attributes in the wild. In Proceedings of the IEEE International Conference on Computer Vision (pp. 3730-3738).
\bibitem {fdl2}
Parkhi, O.M.,\textit{et al.}, 2015, September. Deep Face Recognition. In BMVC (Vol. 1, No. 3, p. 6).
\bibitem {fdl3}
Farfade, S.S., \textit{et al.}, 2015, June. Multi-view face detection using deep convolutional neural networks. In Proceedings of the 5th ACM on International Conference on Multimedia Retrieval (pp. 643-650). ACM.
\bibitem {pe1}
Ranjan, R.,\textit{et al.}, 2017. Hyperface: A deep multi-task learning framework for face detection, landmark localization, pose estimation, and gender recognition. IEEE Transactions on Pattern Analysis and Machine Intelligence.
\bibitem{ed1}
Kahou, S.E., Pal, C.,\textit{et al.}, 2013, December. Combining modality specific deep neural networks for emotion recognition in video. In Proceedings of the 15th ACM on International conference on multimodal interaction (pp. 543-550). ACM.
\bibitem{ijba}
Klare, B.F.,\textit{et al.}, 2015. Pushing the frontiers of unconstrained face detection and recognition: IARPA Janus Benchmark A. In Proceedings of the IEEE Conference on Computer Vision and Pattern Recognition (pp. 1931-1939).
\bibitem{fddb}
Jain, V. and Learned-Miller, E., 2010. Fddb: A benchmark for face detection in unconstrained settings. University of Massachusetts, Amherst, Tech. Rep. UM-CS-2010-009, 2(7), p.8.
\bibitem{wider}
Yang, S.,\textit{et al.}, 2016. Wider face: A face detection benchmark. In Proceedings of the IEEE Conference on Computer Vision and Pattern Recognition (pp. 5525-5533).
\bibitem{rcnn}
Zhu, C.,\textit{et al.}, 2017. CMS-RCNN: contextual multi-scale region-based CNN for unconstrained face detection. In Deep Learning for Biometrics (pp. 57-79). Springer, Cham.
\bibitem{fastrcnn}
Girshick, R., 2015. Fast R-CNN. In Computer Vision (ICCV),  IEEE International Conference on. pp. 1440-1448 IEEE.
\bibitem{fasterrcnn}
Ren, S., \textit{et al.}, 2015. Faster r-cnn: Towards real-time object detection with region proposal networks. In Advances in neural information processing systems (pp. 91-99).
\bibitem{facefaster}
Jiang, H. and Learned-Miller, E., 2017, May. Face detection with the faster R-CNN. In Automatic Face \& Gesture Recognition (FG 2017), 2017 12th IEEE International Conference on (pp. 650-657). IEEE.
\bibitem{ssearch}
Uijlings, J.R., Van De Sande, K.E., Gevers, T. and Smeulders, A.W., 2013. Selective search for object recognition. International journal of computer vision, 104(2), pp.154-171.
\bibitem{vgg16}
Long, J.,\textit{et al.}, 2015. Fully convolutional networks for semantic segmentation. In Proceedings of the IEEE conference on computer vision and pattern recognition (pp. 3431-3440).
\bibitem{zfnet}
Zeiler, M.D. and Fergus, R., 2014, September. Visualizing and understanding convolutional networks. In European conference on computer vision (pp. 818-833). Springer, Cham.
\end{thebibliography}
\end{document}